\documentclass[11pt,a4paper]{article}
\usepackage[hyperref]{acl2020}
\usepackage{times}
\usepackage{latexsym}
\usepackage{graphicx}
\usepackage{multirow}

\usepackage{microtype}

\aclfinalcopy 

\title{Cascaded Models for Better Fine-Grained Named Entity Recognition}

\author{Parul Awasthy \and Taesun Moon \and Jian Ni \and Radu Florian\\
  IBM Research AI \\
  Yorktown Heights, NY 10598 \\
  USA \\
  \texttt{{awasthyp,tmoon,nij,raduf}@us.ibm.com}
}
\date{}

\begin{document}
\maketitle
\begin{abstract}
Named Entity Recognition (NER) is an essential precursor task for many natural language applications, such as relation extraction or event extraction. Much of the NER research has been done on datasets with few classes of entity types (e.g. PER, LOC, ORG, MISC), but many real world applications (disaster relief, complex event extraction, law enforcement) can benefit from a larger NER type set. More recently, datasets were created that have hundreds to thousands of types of entities, sparking new lines of research \cite{sekine-2008-extended,Ling:2012:FER:2900728.2900742,DBLP:journals/corr/GillickLGKH14,choi-etal-2018-ultra}. In this paper we present a cascaded approach to labeling fine-grained NER, applying to a newly released fine-grained NER dataset that was used in the TAC KBP 2019 evaluation \cite{Ji:2019:tac2019}\footnote{Text Analysis Conference, organized by NIST, in the Knowledge Base Population area, Entity Detection and Linking evaluation.}, inspired by the fact that training data is available for some of the coarse labels. Using a combination of transformer networks, we show that performance can be improved by about 20 $F_1$ absolute, as compared with the straightforward model built on the full fine-grained types, and show that, surprisingly, using course-labeled data in three languages leads to an improvement in the English data. 

\end{abstract}

\section{Introduction}

The main goal of Named Entity Recognition (NER) is to identify, in
unstructured text, contiguous typed references to
real-world entities, such as persons, organizations,
facilities, etc. It is useful as a precursor
to identifying semantic relations between
entities (to fill relational databases), and events
(where the entities are the events arguments). Traditional NER work has focused on coarse-grained
entity types, e.g., 4 entity types in
CoNLL’02 data \cite{TjongKimSang:2002:ICS:1118853.1118877}, 7 entity types in ACE’05 data \cite{Walker2006}.

However, many real-world applications (e.g., disaster
relief, technical support, cybersecurity, law enforcement) may require
a larger set of fine-grained entity types to work properly. Obtaining data for this fine grained, larger type system can be expensive and cumbersome. This paper focuses on describing an approach to use pre-existing small type sets to build an NER system with a larger, more fine-grained type set with limited amount of annotated data for the fine-grained type.

Historically, named entity recognition was initally performed on datasets that had a small number of tags, such as in MUC-6 \cite{grishman-sundheim-1995-design}, MUC-7 \cite{chinchor-1998-overview}, CoNLL'02 and '03 evaluations \cite{TjongKimSang:2002:ICS:1118853.1118877, tjong-kim-sang-de-meulder-2003-introduction}, ACE \cite{doddington-etal-2004-automatic}. More recently, starting with \citet{sekine-2008-extended}, and continuing with the work of  \citet{Ling:2012:FER:2900728.2900742}, \citet{DBLP:journals/corr/GillickLGKH14}, and \cite{choi-etal-2018-ultra}, the size of the entity nomenclature size has increased from 120 to over 10,000. In this paper, we will be using a recently released dataset of 187 named entities, created under the DARPA AIDA program, that was used in the NIST-organized Text Analysis Conference Knowledge-Base Population Entity Detection and Linking (TAC-EDL) 2019 task \cite{Ji:2019:tac2019}, described in Section \ref{sec:data}. The entity tags are hierarchical, including a base type (such as \emph{per(son)}, \emph{org(anization)}, etc.), a subtype (e.g. \emph{per.politician}), and a sub-subtype (e.g. \emph{per.politician.head\_of\_government}).

Given the hierarchical nature of the tag set, and the availability of data for the main type (as used in previous TAC KBP-EDL evaluations), we hypothesize that a hierarchical model that uses a standard NER model to detect the main type (in our case, a BERT-based model), followed by a instance-based mention instance classifier for fine-grained NER (based on a RoBERTa architecture) will perform better than a straight-forward NER model. We show in Section \ref{sec:experiments} that this is indeed the case, and that such a model improves the performance significantly, from 42 F$_1$ to 61.6 F$_1$ . 

\section{Prior Work}
\label{sec:prior}

Named entity recognition (NER) is a subfield of
NLP with a long, established history and a vast literature.
For a good overview of the problem and
the main corpora involved from a classical perspective,
see \cite{ner-sekine2007}. \citet{hammerton-2003-named} was one of the earliest attempts at
using a neural network (specifically, an LSTM) for
NER, though with performance marginally above
baseline. \citet{Collobert:2011:NLP:1953048.2078186} and \citet{lample-etal-2016-neural} were more successful and influential approaches to using neural networks for NER.

Pretrained word-embeddings called
word2vec \citet{Mikolov:2013} proved critical in helping neural architectures achieve
state-of-the-art results across a variety of tasks in NLP including NER \cite{lample-etal-2016-neural}.
The development of contextual or context aware
pretrained word-embeddings such as ELMo \cite{DBLP:journals/corr/abs-1802-05365} and BERT \cite{BERT18}
pushed the SotA frontier for virtually all NLP
tasks even further. For example, merely using
BERT with a final feed-forward layer and
supervised fine-tuning achieved SOTA for a
short-period in English CoNLL NER.

There is a wide and disparate literature for information
extraction when supervised data is either
restricted or unavailable and the type system
is either large, unbounded or unspecified. \citet{Banko:2007:OIE:1625275.1625705} introduces the idea of open information extraction,
where relation triples are discovered from
the general web without any labeled data. \citet{cimiano_volker2005} is related in spirit to this
study in that it attempts to classify named entities
according to a large ontology with no training examples.
Theirs is a fully unsupervised approach
using a classical vector space model where they
assign a label from the ontology to a named entity
by measuring similarity between the label and
named entities through a context vector computed
from a large corpus. \citet{Evans03aframework} is a
thoughtful attempt at the even harder problem of
deriving a type system from scratch from an unlabeled
corpus by using a set of heuristics. \citet{Brambilla2017ExtractingEK} attempts to detect emerging entities in social
media.
\citet{Ling:2012:FER:2900728.2900742} introduced FIGER, a fine-grained NER system built with 112 types, treating the problem as multi-label, multi-class classification with data that was automatically extracted. \citet{mai-etal-2018-empirical} show experiments with fg-NER, a fine grained NER system based on the dataset of \citet{sekine-2008-extended}, which has 200 types. The authors empirically investigate both classical (rule-based, dictionary-based) approaches, CRF+SVM approaches, and LSTM+CNN+CRF based approaches  \cite{ma-hovy-2016-end}, concluding that the latter works well for English but not as well for Japanese, where they remove the CNN layer
and add dictionary and category embeddings to improve performance significantly.

\citet{choi-etal-2018-ultra} introduces a much larger dataset of types, 10201 ultra fine-grained types, also labeled with the \cite{Ling:2012:FER:2900728.2900742} 120 fine-grained types, and 9 general types; the data was annotated using crowd-sourcing. The model prediction is done mention by mention, representing the mention context by adding indices corresponding to whether a word is before, inside, or after the mention, and use and LSTM to represent the context of the mention, then classify the type of the mention based on this representation, and also add a distantly supervised multi-task objective. This method is reminiscent on the mention classification in Section \ref{sec:fine-grained-md}, where we use a RoBERTa encoder instead of a biLSTM model, and we encode the mention differently.

\section{Data\label{sec:data}}
Though, there are a several fine-grained entity type systems proposed for NER research, like FIGER \cite{Ling:2012:FER:2900728.2900742}, YAGO \cite{Suchanek:2007:YCS:1242572.1242667}, in this paper we work on the AIDA NER type system defined for the DARPA AIDA program. This type system contains about 187 3-level hierarchical tags in the form of type.subtype.subsubtype, where subsubtype or subtype.subsubtype could be undefined, given the context. E.g. "per", "per.politician", "per.politician.governor" are all valid tag types. This dataset was used for the TAC2019 KBP-Entity Data Linking Task \cite{Ji:2019:tac2019} where the goal was to produce a type at the finest-grained level that can be confidently labeled, backing off to a higher level if the context did not support the fine-grained labeling. The task mentions that \begin{quote} In an AIDA scenario, there is much informational conflict about the entities that participate in crucial newsworthy events and relations, and fine-grained entity types are needed to represent different hypotheses about the scenario (e.g., is the shooter in an attack event a Military Personnel, or a Demonstrator?)
\end{quote}.  
There are in about 118 community annotated documents available for this task\footnote{The authors are very thankful to Heng Ji and her team at UIUC, who organized the annotation and shared the data with the community  - this research would not have been possible otherwise.}, and an additional 65 documents that were released for AIDA project. These 65 documents contain partial annotations, i.e. not all mentions of names are labeled in this data. Additionally 10 documents were released as the feedback documents in the course of the task. The evaluation set comprises of 394 documents.

We use the community annotated data as the training set, the AIDA and feedback documents as development data, and 394 evaluation documents as the test. The numbers on this paper are reported on the test set.

The evaluation data, and our model, focuses on \textbf{Named} mentions only. Nominals, pronouns and non-named mentions are excluded.

Given the lack of actual training data provided specifically for this Fine-Grained type system, we have investigated the previously
released datasets – specifically ACE’05 \cite{Walker2006} and
TAC’17. The TAC17 dataset was more recently
released, but had fewer types (as it does not include
VEHICLE and WEAPON), while at the
same time matches more closely the annotation
guidelines. In the end, after initial experimentation,
we have decided to augment the TAC'17 dataset with silver-labeled mentions by running an ACE’05 trained
system on the train/development sets, and adding any mentions of the type WEAPON and VEHICLE that do not overlap with the gold annotations. In addition, we have also used an in-house
mention detection system to add two other types -
COM and LAW - by running the training and development
set through an in-house built classifier [citation removed for anonymity] 
and retaining only mentions of the two types that do not
overlap with any existing annotations.

\section{Large Space Mention Detection}
In order to balance the large number of entity types
(187) with the need for sufficient training data to
perform sequential tagging for mention detection,
we have decided to solve the problem with the following
two steps:

\begin{itemize}
\item First, perform coarse-grained mention detection
in the usual fashion, by converting it
to an IOB sequential token prediction task
\cite{sang99representing} – the
type system we used here is composed of 9
entity types (CRM, FAC, GPE, LAW, LOC,
ORG, PER, VEH, WEA).

\item Second, classify each mention obtained in the
previous step with a subtype.subsubtype label
- an example-based classification task that
takes the full sentence of the mention into account.
\end{itemize}

This way the first step focuses on producing the mention boundary and the second step on producing the type for the mention. Below we describe the salient characteristics of these steps.

\subsection{Coarse-Grained Mention Detection}
The coarse-grained mention detector is a BERT based
mention detector, as described below.

\begin{figure}
\begin{centering}
\vspace*{-4mm}
\hspace*{-4mm}
\includegraphics[scale=0.37]{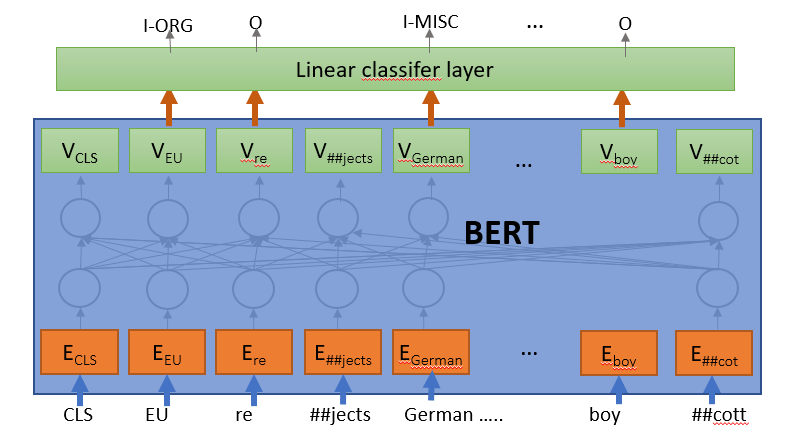}
\par\end{centering}
\caption{BERT-ML Architecture\label{figure:BERT}}
\vspace*{-5mm}
\end{figure}

We approach NER in a standard fashion: a sequence labelling task which assigns a tag to each word based on its context. Given a sentence $\left\{w_{1},w_{2},....w_{n}\right\}$, we feed it to the BERT model to obtain contextual BERT embeddings for each word as \{$v_{1},v_{2},...v_{n}$\}, capturing each word's context
via many attention heads in each of its layer. These embeddings are then fed to a linear feed forward 
layer to obtain labels \{$y_{1},y_{2},...y_{n}$\} corresponding to each each word (see Figure \ref{figure:BERT}). This entire network is trained with each epoch thereby fine-tuning the BERT embeddings for the NER task. We are using an IOB2 encoding 
of the entities \cite{sang99representing}, as it performed best in preliminary results.

We use the HuggingFace PyTorch implementation of Transformers \cite{Wolf2019HuggingFacesTS} and the
BERT WordPiece Tokenizer. We follow the recipe proposed by \citet{BERT18} for building
named entity taggers: to convert the NER tags from tokens to word pieces, we assign the tag of the token to
its first piece, then assign the special tag 'X' to all other pieces. No prediction is made for $X$ tokens during training and testing. Figure \ref{figure:BERT} shows both the architecture of the proposed model, and the NER annotation style.

\subsection{Fine-Grained Mention Detection \label{sec:fine-grained-md}}

As not much gold data is available for the fine-
grained mention detection task, we approach
this step as a standard
example based classification task that assigns a tag to each mention based on its context. Given
a sentence and the mention boundary in the sentence as input,
we consider the task of classifying the example with a fine-grained type of form $type.subtype.
subsubtype$. Details are described below:

Training takes as input:
\begin{enumerate}
    \item a sentence in the form of $\left\{w_1,w_2,...w_n\right\}$, with k mentions at
$\left\{(w_{i_1} : w_{j_1}), (w_{i_2} : w_{j_2}), .... (w_{i_k} : w_{j_k})\right\}$, where the
mention span $(w_{i_x} : w_{j_x})$ begins
at token position $i_x$ and ends at $j_x$; 
    \item a coarse-grained label (type only) $z_x$ for each of the mentions. 
    \item a fine-grained (type.subtype.subsubtype) label $y_x$ for each of
these mentions;
\end{enumerate}
\noindent We create $k$
examples from the sentence with a special representation
$W_x$ for $w_{i_x} : w_{j_x}$ in the $x_{th}$ example,
thereby producing the below representations for the
example:

$\left\{w_1,w_2,\ldots,w_{{i_1}-1},W_1,w_{{j_1}+1},\ldots,w_n\right\}$

$\left\{w_1,w_2,\ldots,w_{i_2-1},W_2,w_{j_2+1},\ldots,w_n\right\}$

\ldots

$\left\{w_1,w_2,\ldots,w_{i_k-1},W_k,w_{j_k+1},\ldots,w_n\right\}$
\begin{figure}
\begin{centering}
\vspace*{-4mm}
\hspace*{-4mm}
\includegraphics[scale=0.21]{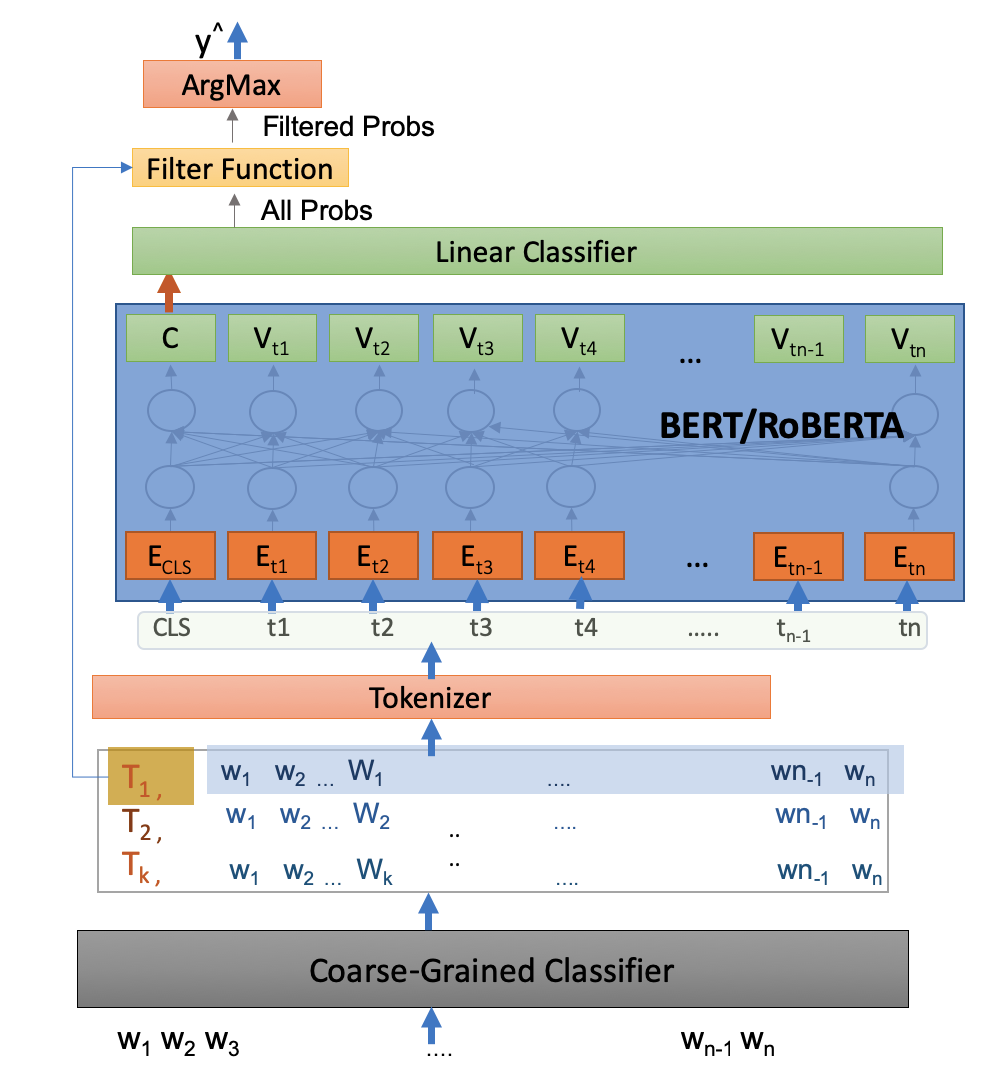}
\par\end{centering}
\caption{BERT-ML Architecture\label{figure:FGArch}}
\vspace*{-5mm}
\end{figure}

These examples are then fed to a transformer based model to obtain a contextual representation
for each sentence. This representation is fed to a
linear layer to get probabilities for each possible fine-grained
class. The system diagram is shown in Figure \ref{figure:FGArch}.

The system is trained using a standard cross-entropy loss over the output probabilities.
But at decoding time we also try ways to restrict the probabilities output space as follows.

Let the $\left\{p_{i_0} , p_{i_1}....p_{i_l}\right\}$ be the representation of
the probabilities produced by the classifier for example
i where $p_{i_k}$ is the probability for label k.
Then we produce $y_i$ as:

\[y_i = \arg\max(F(\left\{p_{i_0}, p_{i_1},...p_{i_l}\right\}, z_i)) \]

\noindent $F(\cdot)$ is a function that given $l$ probabilities and a
coarse label $z_i$ filters out some of the $l$ probabilities
and produces an array of probabilities with
size less than or equal to $l$ so that the possible output
space is reduced\footnote{Note that this approach is equivalent to setting to $0$ all the probabilities that we want to 'skip' and renormalizing among the non-$0$ values}. We try different
representations for $W_i$ ans $F(\cdot)$ in our experiments.
These variations are described below.

\subsubsection{Mention Representations}
Here we describe the representations we tried for
the mention text ($W$).

\vspace*{1mm}\noindent\textbf{Masking the mention}: We replace the mention tokens
in the sentence with a tokenizer specific mask
token, and then add the mention token to the end
of the sentence after a tokenizer specific separator, as exemplified below
\begin{quote}
    \textit{\textbf{Alice} was beginning to get very tired of sitting by her sister on the bank .}

$<$\textbf{MASK}$>$ was beginning to get very tired of
sitting by her sister on the bank . $<$\textbf{SEP}$>$ \textbf{Alice}

\end{quote}

\if false
\noindent\textbf{Masking with Coarse-Grained Type}: We replace
the mention tokens in the sentence with a
the special token, the coarse-grained mention type of
the mention. To do this we add all the coarse
grained types to the tokenizer vocabulary, so they
are not split, as below
\begin{quote}
    \textit{\textbf{Alice} was beginning to get very tired of sitting
by her sister on the bank .}

\textbf{PER} was beginning to get very tired
of sitting by her sister on the bank .
\end{quote}
\fi
\noindent\textbf{Bounding with Coarse-Grained Type}: We surround
the mention tokens with the the coarse-grained mention type of
the mention. Again, we add all the coarse
grained types to the tokenizer vocabulary, so they
are not split. E.g.
\begin{quote}
    \textit{\textbf{Alice} was beginning to get very tired of sitting
by her sister on the bank .}

\textbf{PER Alice PER} was beginning to get very tired
of sitting by her sister on the bank .
\end{quote}
\subsubsection{Filter Functions}
To restrict the label space to the most likely,
while decoding, not while training, we try a few different
filter functions. These functions use the coarse-grained label and the probabilities produced by the classifier to narrow down the output space of the labels. 

As mentioned earlier, for this task each class is represented
as \textit{type.subtype.subsubtype}. Let the input to
this function be a set of probabilities $p_1, p_2...p_l$
where pi is the probability of the $i_{th}$ class, and a
coarse-grained type $t$. The output is again a set of
probabilities $\left\{p_x, p_y,...p_z\right\}$ such that $|$output$|\le l$.

\vspace*{3mm}\noindent\textbf{Pass-through Filter Function} returns
the same input it received with no filtering.

\[ F(\left\{p_1,...p_l \right\} ,t) = \left\{p_1,...p_l\right\}\]

\vspace*{3mm}\noindent\textbf{Coarse-Grained Type based Filter Function}  returns probabilities of only those classes
for which the type part of the class is same as
coarse-grained type. All other probabilities are filtered
out. E.g. if coarse-grain type for a mentions
is PER then the output of $F()$ would be only classes
with PER type and no classes with other types line
GPE, FAC,etc. irrespective of their probabilities.

\[F(\left\{p_1,...p_l\right\}, t) = \left\{p_x\textnormal{ s.t. }E(x)=t \right\}\]
\noindent where $E(x)$ is the main type of the label $x$ (e.g. $E($per.politician.senator$)=\textnormal{per}$).

\vspace*{3mm}\noindent\textbf{Threshold based Filter Function} is
similar to the above Coarse-Grained Type based
filter with a small difference. It still returns probabilities
of those classes for which the type part
of the class is same as coarse-grained type, but it
also returns probabilities of classes of other types,
if the in probability of that class is greater than a threshold $\theta$. By doing this the model is able to switch types if it is really sure of the type, even when it does not match the coarse-grained type.

\[
F(\left\{p_1,...p_l\right\}, t, \theta) =
\left\{p_x \textnormal{ s.t. }E(x)=t \vee p_x\ge\theta \right\}
\]

\if false 
We use Huggingface pytorch Transformers for
the fine grained annotation task. We train the
model on the community annotated data shared by
UIUC and we test on AIDA phase 1 Eval data. At
training time we train on gold coarse-grained labels
but at decode time we use the system coarse
grained labels produced by the NER model.
\fi

\subsection{Advantage of Cascaded Models}

One advantage of cascaded models is worth mentioning here. As noted earlier, some of the labeled data we had access to (the AIDA-labeled data) is only partially annotated, and a token sequence classifier would have issues with that style of annotation, as the mentions that are not annotated would be interpreted by the system as not entities. However, the cascaded system, in particular the fine-grained classification system can incorporate those examples straightforwardly, as it is trained on positive examples only.

\section{Experiments \label{sec:experiments}}

\subsection{Experimental Setup}

We try several transformer-based pre-trained models, including \textit{bert-base-uncased, bert-large-cased,  bert-large-cased-whole-word-masking, bert-base-multilingual-cased, roberta-large}, etc. 
\begin{table*}
\
\begin{center}
 \begin{tabular}{|c||c| c|c|} 
 \hline
 Mention Representation & Acc-SST&Acc-ST&Acc-T \\ 
 \hline
 \small{Masked Mention} & 67.63&75.11 & 90.08\\ 
 \hline
 \small{Entity-Masked Mention} & 67.91& 75.76&92.56\\
 \hline
 \small{Entity-Bounded Mention} & 68.81&77.09&94.15\\
 \hline
\end{tabular}
\end{center}
\caption{Accuracy of the Fine-Grained models across different mention representations at various hierarchical levels } \label{table:fg-accuracy}
\end{table*}

Interestingly, the TAC17 dataset has annotated coarse labels in three languages: English, Spanish, and Chinese. Even though the TAC19 evaluation is run only on English, given the availability of the datasets and the emerging multilingual NLP approaches \cite{Wo-Dredze-2019-beto, Pires-etal-2019-multilingual-bert, moon2019lingua} we also used the three languages combined to produce a multilingual BERT system, which we used to label the test set with coarse labels. The obvious advantage here is that we have more data (the English dataset has about 540k tokens, while all three languages combined have 1.4M tokens), with the disadvantage that the model has fewer parameters, as we had to use the bert-base-multilingual-cased pretrained model (as opposed to the bert-large-uncased or bert-large-cased-whole-word-masking pretrained models).

For Coarse-Grained NER we train with a learning rate of 1e-5, with a batch size of 64 for 20 epochs.

For Fine-Grained classification we try several learning rates in the range of 2e-5 to 5e-5 with a batch size of 24 and train for 10 epochs.

In both cases we select models that perform the best on the development set.

Here we report numbers on the TAC KBP:EDL 2019 evaluation corpus that comprises of 394 documents.

\vspace*{3mm}\noindent\textbf{Baseline Model}
To compare our approach we train a baseline NER model that is trained end-to-end with the same training and development data as used by our fine-grained classification model - a standard BERT model. Again we choose the model that performs the best on the development data and evaluate it on the test. 

\subsection{Results}

\begin{table}
\begin{center}
 \begin{tabular}{|c|c|c|c|} 
 \hline
 \small{BERT Model} & \small{P} & \small{R} & \small{F$_1$} \\ 
 \hline
 \hline
 \small{large-uncased} & \small{77.3\tiny{$\pm$0.8}} & \small{77.2\tiny{$\pm$2.1}} & \small{77.3 \tiny{$\pm$1.0}} \\
 \hline
 \small{large-case-wwm} & \small{78.1\tiny{$\pm$1.4}} & \small{78.9\tiny{$\pm$1.9}} & \small{78.5\tiny{$\pm$1.3}} \\
 \hline
 \small{base-multi-case} & \small{78.8\tiny{$\pm$0.3}} & \small{80.8\tiny{$\pm$0.2}} & \small{79.8\tiny{$\pm$0.2}} \\
 \hline
\end{tabular}
\end{center}
\caption{Performance of Coarse grained models, averaged over 5 runs with random seeds. Interestingly, not only is the multilingual model better than the large English model, but it is also more consistent, as can be seen by the low standard deviation in F$_1$ scores.} \label{table:Coarse-grained-results}
\end{table}

Scores on the 9 high-level types (CRM, FAC, GPE, LAW, LOC,
ORG, PER, VEH, WEA) produced by two of the highest scoring Coarse-Grained models are shown in Table \ref{table:Coarse-grained-results}. Here we measure only the $type$ part of the $type.subtype.subsubtype$ label. The interesting point to note is that the bert-base-multilingual architecture is able to produce numbers that are better than the bert-large-cased (whole-word-masking, wwm) architecture, in addition to having much lower variance over 5 models trained with same hyper-parameters (but with different seeds): the standard deviation is 0.2 for multilingual model, and ~1.3 for bert-large English only models.

The results for fine-grained mention detection
are given in Table \ref{table:fg-accuracy}. We report the accuracy numbers using gold coarse-grained types in this table. These models use the \emph{Pass-through Filter} and so chooses from all possible labels without any restrictions. The numbers are shown across the $type$ (T), $type.subtype$ (ST) and $type.subtype.subsubtype$ (SST) levels. So if the gold label for a mention is \emph{per.politician.governor} and the system produces {per.politician} then the system would get a score of 1 at the T and ST level but would get 0 for the SST level. Looking at the data at these three levels is useful, as in real-world cases, this backed-off label might be more useful than not producing a label.  \if We see that the Entity bounded mention provides the best numbers here. We posit that bounding a mention in place with the coarse type helps in preserving the flow of the sentence.In the other two cases, the mention text is masked using some special tokens, and the original text is added to the end of the sentence, changing the sentence structure.\fi

\begin{table*}
\begin{center}
 \begin{tabular}{|c|c|c| c|c|} 
 \hline
 Coarse NER Model & Mention Reps & P&R&F \\ 
 \hline
 \multicolumn{2}{|l|}{\hspace{8mm}Baseline}&35.53&  51.20&  41.95\\
 \hline
 \multirow{2}{*}{bert-large-cwwm}&Masked-Mention&60.50&62.14&61.31\\
 \cline{2-5}
    &Entity-Bounded Mention&60.47&62.11&61.28\\
 \hline
 \multirow{2}{*}{bert-base-multi}&Masked-Mention&60.7&62.43&61.60\\
 \cline{2-5}
    &Entity-Bounded Mention&58.39&59.96&59.16\\
 \hline
\end{tabular}
\end{center}
\caption{Performance of baseline vs two-step system models} \label{table:fg-e2e}
\end{table*}

Table \ref{table:fg-e2e} shows the performance of the various models across the whole task. The baseline model is an end-to-end NER model, that does both boundary detection and mention classification at the same time. As can be seen, our two-step approach does ~19 points better than the baseline. The numbers from our models are again using the pass-through filter, where the model can choose the mention type with the highest probability, no matter what is the coarse-grained type. Most of our models perform comparably, with no clear winner across the mention representation approach. We discuss possible reasons for this in the error analysis in Section \ref{sec:discussion}.

\begin{table*}
\begin{center}
 \begin{tabular}{|c|c|c| c|c|} 
 \hline
 Model & Filter Type & P&R&F \\ 
  \hline
 \multirow{3}{*}{Masked-Mention}&Pass-through& 60.50&62.14&61.31\\
 \cline{2-5}
    &Coarse-Grain Type&59.09&60.70&59.88\\
    \cline{2-5}
    &Threshold Based&60.09&61.72&60.89\\
 \hline
 \multirow{3}{*}{Entity-Bounded Mention}&Pass-through&60.47&62.11&61.28\\
 \cline{2-5}
    &Coarse-Grain Type&59.01&60.61&59.80\\
    \cline{2-5}
    &Threshold Based&60.24&61.88&61.05\\
 \hline
\end{tabular}
\end{center}
\caption{Performance of the full model across Filter Types} \label{table:fg-filter}
\end{table*}

Finally we show the results with the other filtering strategies in Table \ref{table:fg-filter}. The threshold for threshold based approach is set as 0.9 as that showed the best results on development data. As can be seen the Pass-through strategy, that lets the classification model learn the label type, and discard the coarse-grained type, does better in both mention representations. This ties back to our observation about the Mention Representations, and we discuss possible reasons for this in the error analysis in Section \ref{sec:discussion}.

\if
We also investigated building ensemble systems,
but the performance improvements we obtained
with voting was minimal (less than .2F).
The voting scheme is a good approach for hedging
against bad models (models that over-train on
the development dataset), as the combined model
will tend to be more robust, but the large data size
(300,000 documents) that needed to be processed
during the evaluation was a concern, so we decided
to use the best performing model from the
development set, instead of an ensemble classifier
for the final evaluation.
\fi

\section{Discussion\label{sec:discussion}}
As seen in Table \ref{table:fg-e2e} our model outperforms the baseline by almost 20 F$_1$. After investigating the system output, a few explanations crystallized:
\begin{itemize}
    \item More data: As we can see the multilingual model, despite being a smaller-sized bert-base model, performs competitively with the bert-large model. It has been shown in literature \cite{Wo-Dredze-2019-beto, Pires-etal-2019-multilingual-bert, moon2019lingua} that adding more data, even in a different language, can help NER, and this is shown here, even for English.
    \item Two-step approach: By breaking the task into a two step approach, we benefit from using a large labelled corpus for mention boundary detection and use the smaller corpus for mention classification. The numbers for this are shown in \ref{table:boundary}. The TAC'17 data based model does more than 20 points better on boundary detection.
    \item The two step approach is able to use partially annotated data for the fine-grained classification, which is positive, given that some of the available data is only partially labeled.
\end{itemize}

\begin{table}
\begin{center}
 \begin{tabular}{|c|c|c|c|} 
 \hline
 \small{Model} & \small{P} & \small{R} & \small{F$_1$} \\ 
 \hline
 \hline
 \small{baseline} & \small{60.02} & \small{81.44} & \small{69.11} \\
 \hline
 \small{base-multi-case} & \small{88.49} & \small{90.88} & \small{89.67} \\
 \hline
\end{tabular}
\end{center}
\caption{Performance of Baseline vs. Two-Step model on boundary detection} \label{table:boundary}
\end{table}

Another  observation one can make from Table \ref{table:fg-filter} is that the \emph{Pass-Through} strategy performs better than strategies that use the type produced by the coarse-grained model, which is rather surprising, given the apparent simplicity of the pass-through model. In addition, the mention representation that uses the coarse-grained type does not perform any better than the one using $<$MASK$>$ token to create sentence representation. 

Investigating these observations, we noted that the annotation guidelines for TAC2019 EDL task are different from the past tasks like TAC'17. In particular, the entities \emph{UN} and \emph{NATO} are labeled as \textbf{ORG} in TAC'17 and previous tasks, but for TAC2019 EDL, these are labeled as \textbf{GPE}, specifically, \emph{gpe.organizationofcountries.organizationofcountries}. These 2 tokens are very frequent, having over 100 examples. This renders the type produced by the coarse-grained type as unhelpful, and end up confusing the model: the fine-grained model does better learning to remap the type from the fine-grained examples in the TAC2019 EDL data, rather than sticking to the coarse-grained types. As shown in Table \ref{table:fg-filter}, the system is able to effectively convert the types to the appropriate ones, resulting in improved performance.

\if
As the community annotated training data contains
examples for only 122 types, our model can
make predictions only for those types and always
misses the other types.
\fi

\section{Conclusion}
We present in this paper a method of performing fine-grained named entity recognition in two
stages: first perform coarse-grained NER using a BERT-based token classification model, followed
by an instance-level fine-grained classification.
The model is trained on both English and multilingual data reused from previous evaluations (TAC’17), augmented
with labels from ACE’05 and an in-house dataset
to add four types - vehicle, weapon,
crime, and law. Coarse-grained models are
based on a BERT (large uncased) model and the BERT base multilingual model, while
the course-grained model is based on a RoBERTa
(large cased) model. Interestingly, the multilingual model (even if has fewer parameters) performs
 better than the English model, and results in a small improvement in performance, but is a lot more stable (as shown by the small variance in performance among 5 models trained with different random seeds). 
 
 The best combination of parameters and system choices outperforms the straightforward model of building a sequence classifier on the fine-grained data by almost 20 $F_1$, showing the effectiveness of the method.


\bibliography{anthology,acl2020,jian,emnlp-ijcnlp-2019,sil,acl20}
\bibliographystyle{acl_natbib}

\end{document}